# Collective Intelligence, Data Routing and Braess' Paradox

**David H. Wolpert**                                    DHW@PTOLEMY.ARC.NASA.GOV
*NASA Ames Research Center, Mailstop 269-2*
*Moffett Field, CA 94035*

**Kagan Tumer**                                         KAGAN@PTOLEMY.ARC.NASA.GOV
*NASA Ames Research Center, Mailstop 269-3*
*Moffett Field, CA 94035*

## Abstract

We consider the problem of designing the the utility functions of the utility-maximizing agents in a multi-agent system (MAS) so that they work synergistically to maximize a global utility. The particular problem domain we explore is the control of network routing by placing agents on all the routers in the network. Conventional approaches to this task have the agents all use the Ideal Shortest Path routing Algorithm (ISPA). We demonstrate that in many cases, due to the side-effects of one agent's actions on another agent's performance, having agents use ISPA's is suboptimal as far as global aggregate cost is concerned, even when they are only used to route infinitesimally small amounts of traffic. The utility functions of the individual agents are not "aligned" with the global utility, intuitively speaking. As a particular example of this we present an instance of Braess' paradox in which adding new links to a network whose agents all use the ISPA results in a *decrease* in overall throughput. We also demonstrate that load-balancing, in which the agents' decisions are collectively made to optimize the global cost incurred by all traffic *currently* being routed, is suboptimal as far as global cost *averaged across time* is concerned. This is also due to "side-effects", in this case of current routing decision on future traffic. The mathematics of Collective Intelligence (COIN) is concerned precisely with the issue of avoiding such deleterious side-effects in multi-agent systems, both over time and space. We present key concepts from that mathematics and use them to derive an algorithm whose ideal version should have better performance than that of having all agents use the ISPA, even in the infinitesimal limit. We present experiments verifying this, and also showing that a machine-learning-based version of this COIN algorithm in which costs are only imprecisely estimated via empirical means (a version potentially applicable in the real world) also outperforms the ISPA, despite having access to less information than does the ISPA. In particular, this COIN algorithm almost always avoids Braess' paradox.

## 1. Introduction

There is a long history of AI research on the design of distributed computational systems, stretching from Distributed AI (Huhns, 1987) through current work on multi-agent systems (MAS's) (Claus & Boutilier, 1998; Hu & Wellman, 1998a; Jennings, Sycara, & Wooldridge, 1998; Sandholm, Larson, Anderson, Shehory, & Tohme, 1998; Sycara, 1998). When the individual agents in such a system each have personal utility functions they are trying to maximize and we also have a 'world utility' that rates the possible dynamic histories of the overall system, such a MAS constitutes a 'collective'. In this paper we are particularly concerned with agents that use machine learning techniques (e.g., Reinforcement Learning (RL) Kaelbing, Littman, & Moore, 1996; Sutton & Barto, 1998; Sutton, 1988; Watkins & Dayan, 1992) to try to maximize their utilities.

The field of Collective Intelligence (COIN) is concerned with the central design problem for collectives (Wolpert, Tumer, & Frank, 1999; Wolpert & Tumer, 1999): *How, without any detailed modeling of the overall system, can one set utility functions for the individual agents in a COIN so that the overall dynamics reliably and robustly achieves large values of the provided world utility?* In other words, how can we leverage an assumption that our learners are individually fairly good at what they do, to have the collective as a whole perform well? [1]

An example of where this question looms very large is the problem of how to optimize the flow of certain entities (e.g., information packets, cars) from sources to destinations across a network of routing nodes. Here we are concerned with the version of the problem in which "optimization" consists of minimizing aggregate cost incurred by the entities flowing to their destinations, and where an agent controls the routing decisions of each node in the network. This problem underlies the distributed control of a large array of real-world domains, including internet routing, voice/video communication, traffic flows, etc. From the COIN perspective, the problem reduces to the question of what goals one ought to provide to each router's agent so that each agent's self-interestedly pursuing its own utility results in maximal throughput of the entire system ("incentive engineering").

In this paper we investigate the application of recently developed COIN techniques, to this routing domain. Like all work concerning COINs, these techniques are designed to be very broadly applicable, and in particular are not designed for the routing domain. Accordingly, their performance in this domain serves as a good preliminary indication of their more general usefulness.

To ground the discussion, we will concentrate on the telecommunications data routing problem where the entities being routed are packets. Currently, many real-world algorithms for this problem are based on the Shortest Path Algorithm (SPA). In this algorithm each routing node in the network is controlled by an agent who maintains a "routing table" of the "shortest paths" (i.e., sequences of links having minimal total incurred costs) from its node to each of the possible destination nodes in the net. Then at each moment the agent satisfies any routing requests for a particular destination node by sending all its packets down the associated shortest path. Many Ideal SPA (ISPA) algorithms exist for efficiently computing the shortest path when agent-to-agent path-cost communication is available and the costs for traversing each agent's node are unvarying in time, e.g., Dijkstra's Algorithm (Ahuja, Magnanti, & Orlin, 1993; Bertsekas & Gallager, 1992; Deo & Pang, 1984; Dijkstra, 1959).

If a non-infinitesimal amount of traffic is to be routed to a particular destination at some moment by some agent, then that agent's sending all that traffic down a single path will not result in minimal cost, no matter how that single path is chosen. However if it must choose a single path for all its traffic, and if the routing decisions by all other agents are fixed, then tautologically by using the ISPA the agent chooses the best such path, as far as the traffic it is routing is concerned. Accordingly, in the limit of routing an infinitesimally

---

small amount of traffic, with all other agents' strategies being a "background", the ISPA is the optimal (least aggregate incurred cost) routing strategy *for the traffic of the associated single agent considered individually.*

One might hope that more generally, if the agent must allot all of its traffic to a single path and all other agents' traffic decisions are fixed, then its choosing that path via the ISPA would be the choice that minimizes total incurred cost of all traffic across the net, at least in the limit of infinitesimally little traffic. This is not the case though, because in using the SPA the agent is not concerned with the deleterious side-effects of its actions on the costs to the traffic routed by other agents (Korilis, Lazar, & Orda, 1997a; Wolpert et al., 1999). The problem is made all the worse if the other agents are allowed to change their decisions in response to our agent's decision. In the extreme case, as elaborated below, if all agents were to try to minimize their personal costs via ISPA's, then the agents would actually *all* receive higher cost than would be the case under an alternative set of strategies. This is an instance of the famous Tragedy Of the Commons (TOC) (Hardin, 1968).

Deleterious side-effects need not be restricted to extend over space; they can also extend over time. Indeed, consider the algorithm of having all agents at a given moment make routing decisions that optimize global cost incurred by the traffic *currently being routed*, an algorithm often called "load-balancing" (LB) (Heusse, Snyers, Guerin, & Kuntz, 1998). By definition, LB avoids the deleterious side-effects over space that can result in the TOC for the costs incurred by the traffic currently being routed. However, due to side-effects over time, even conventional LB can be suboptimal as far as global cost averaged across time is concerned. Intuitively, one would have to use "load-balancing over time" to ensure truly optimal performance. So even if one could somehow construct a distributed protocol governing the agents that caused them to implement LB, still one would not have gotten theme to all act in a perfectly coordinated fashion. Such difficulties make this an appropriate domain in which to investigate how well COIN techniques work in practice.

Real-world SPA's (RSPA) work by applying an ISPA to the estimated costs for traversing each path of every agent. Typically those estimates will be in error because agent-to-agent communication is not instantaneous, and therefore routing tables may be based on out of date information. More generally though, even if that communication were instantaneous, the cost to traverse an agent's node may be different by the time the packet arrives at that node. Accordingly, in general the performance of RSPA's is bounded above by that of the associated ISPA. In this paper we do not wish to investigate such topics, but rather to highlight the issue of side-effects. Accordingly we "rig the game" in our experimental comparisons in favor of the SPA, by using ISPA's rather than RSPA's.

In general, even without side-effects, determining the optimal solution to a flow problem (e.g., determining what the loads on each link need to be to maximize throughput on a non-cooperative data network) can be nontractable (Ahuja et al., 1993; Orda, Rom, & Sidi, 1993b). Therefore, we will concern ourselves with providing *good* solutions that avoid the difficulties the ISPA has with side-effects. It is not our aim here to present algorithms that find the best possible (perfectly load-balanced over time) solution. Previous work on using machine learning to improve routing has sometimes resulted in better performance than (non-idealized) SPA's (Littman & Boyan, 1993; Boyan & Littman, 1994; Stone, 2000; Marbach, Mihatsch, Schulte, & Tsisiklis, 1998). That work has not grappled with the central COIN design problem however.

In Section 2 we discuss SPA's deficiencies and in particular their manifestations in Braess' paradox. Then, in Section 3 we present the theory of collective intelligence, an approach that promises to overcome those deficiencies. We then discuss the routing model we will use in our experiments, and show how the theory of COINs can be applied to that model to provide an alternative to shortest path algorithms in Section 3. In Section 5 we present simulation results with that model comparing ISPA to COINs. These results demonstrate that in networks running ISPA, the per packet costs can be as much as 32 % higher than in networks running algorithms based on COIN theory. In particular, even though it only has access to imprecise estimates of costs (a handicap that does not hold for ISPA), the COIN-based algorithm almost always avoids Braess' paradox, in stark contrast to the ISPA. In that the cost incurred with ISPA's is presumably a lower bound on that of an SPA not privy to instantaneous communication, the implication is that COINs can outperform such real-world SPA's. We conclude that the techniques of the field of collective intelligence can be highly effective in designing the utility functions of the members of a MAS to ensure they work in a coordinated and efficient manner to optimize overall performance.

## 2. Suboptimality of Shortest Path Routing and Braess Paradox

In this section we first demonstrate the suboptimality of an SPA when we have multiple agents making simultaneous routing decisions, where no agent knows ahead of time the other's choice, and therefore does not know ahead of time exactly what the costs will be. We then demonstrate that such suboptimality can hold even when only one agent is making a decision, and it knows what decisions the others have previously made. Next we present Braess' paradox, a particularly pointed instance of these effects (for other discussion of Braess' paradox in SPA routing, see Bass, 1992; Cohen & Kelly, 1990; Cohen & Jeffries, 1997; Hogg, 1995; Glance & Hogg, 1995; Korilis, Lazar, & Orda, 1999).

### 2.1 Suboptimality of SPA

Perhaps the simplest example of how individual greed on the part of all agents can lead to their collective detriment occurs when two agents determine that their shortest path is through a shared link with a limited capacity, while both have a second option that is slightly less preferable. In such a case, their using the common link degrades the performance of *both* parties, since due to limited capacity the performance of that link will quickly fall below that of their second option.

More precisely, consider the case where the shared link has a cost given by $x^3$ when traversed by $x$ packets, and where each router has an optional second link to the destination where the cost for traffic $x$ to traverse such a second link is $2x$. Acting alone, with a single packet to send, they would both send that packet through the shared link (cost of 1). However by both doing so, they incur a larger cost (cost of 8) than if they had both used their second choices (cost of 4). Without knowing what each other will do ahead of time (information not conventionally contained in routing tables), the agents will necessarily have mistaken cost estimates and therefore make incorrect routing decisions. In this, even in the limit of differentially small packets, use of SPA will lead to a wrong routing decision.

## 2.2 Suboptimality of ISPA

We now analyze a situation where the routers may know what the loads are but are each acting to optimize the delays experienced by *their* packets alone. Consider the network shown in Figure 1. Two source routers $X$ and $Y$ each send one packet at a time, with $X$ sending to either intermediate router $A$ or $B$, and $Y$ sending to either $B$ or $C$. This type of network may arise in many different topologies as a subnetwork. Accordingly, difficulties associated with this network can also apply to many more complex topologies.

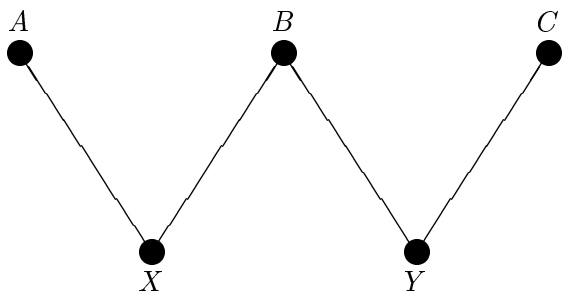

Figure 1: Independent decisions at the source

Let $x_A$, $x_B$, $y_B$, and $y_C$, be the packet quantities at a particular fixed time $t$, at $A$, $B$, or $C$, and originating from $X$ or $Y$, as indicated. At $t$, each source has one packet to send. So each of our variables is binary, with $x_A + x_B = y_B + y_C = 1$. Have $V_i(z_i)$ be the cost, per packet, at the single instant $t$, at router $i$, when the total number of packets at that instant on that router is $z_i$. So the total cost incurred by all packets at the time $t$, $G(\vec{x}, \vec{y})$, equals $x_A V_A(x_A) + (x_B + y_B)V_B(x_B + y_B) + (y_C)V_C(y_C)$.

In an ISPA, $X$ chooses which of $x_A$ or $x_B = 1$ so as to minimize the cost *incurred by $X$'s packet alone*, $g_X(\vec{x}) \equiv x_A V_A(x_A) + x_B V_B(x_B + y_B)$. In doing this the ISPA ignores the $y_B V_B(x_B + y_B)$ term, i.e., it ignores the "side effects" of $X$'s decision. Real-world SPA's typically try to approximate this by having $X$ choose either $A$ or $B$ according to whether $V_A(0)$ or $V_B(y_B)$ is smaller, where those values can be estimated via pings, for example.

The right thing to do from the point of view of minimizing the global cost of course is instead to have $X$ minimize $G(\vec{x}, \vec{y})$, or more precisely, the components of $G(\vec{x}, \vec{y})$ that depend on $X$. Writing it out for this case, $X$ ought to act to minimize $x_A V_A(x_A) + (x_B + y_B)V_B(x_B + y_B)$. Due to the constraint that $x_A + x_B = 1$, this means sending down $A$ iff $V_A(1) < (y_B + 1)V_B(y_B + 1) - y_B V_B(y_B)$, which differs from the ISPA result in that $X$ is concerned with the full cost of going through router $B$, not just the portion of that cost that its packet receives.

In the context of this example, this $G$-minimizing algorithm constitutes "load-balancing" (LB). Note that so long as $\text{sgn}[V_A(0) - V_B(y_B) - y_B V_B'(y_B)] \neq \text{sgn}[V_A(0) - V_B(y_B)]$, even in the limit of infinitesimally small traffic (so that $x_A + x_B$ equals some infinitesimal $\delta$), ISPA and LB still disagree. LB considers side-effects of current routing decisions on other traffic currently being routed. However because it does not consider side-effects of routing decisions on future traffic, even LB may not optimize global cost averaged across all time, depending on the details of the system. However through the use of "effect sets" COINs can account even for such delayed side-effects[2].

## 2.3 Braess' Paradox

Let us conclude this section with an illustration of Braess' paradox (Bass, 1992; Cohen & Kelly, 1990; Cohen & Jeffries, 1997; Glance & Hogg, 1995; Hogg, 1995; Korilis, Lazar, & Orda, 1997b; Korilis et al., 1999), a phenomenon that dramatically underscores the inefficiency of the ISPA. This apparent "paradox" is perhaps best illustrated through a highway traffic example first given by Bass (Bass, 1992): There are two highways connecting towns S and D. The cost associated with traversing either highway (either in terms of tolls, or delays) is $V_1 + V_2$, as illustrated in Net A of Figure 2. So when $x = 1$ (a single traveler) for either path, total accrued cost is 61 units. If on the other hand, six travelers are split equally among the two paths, they will each incur a cost of 83 units to get to their destinations. Now, suppose a new highway is built connecting the two branches, as shown in Net B in Figure 2. Further, note that the cost associated with taking this highway is not particularly high (in fact for any load higher than 1, this highway has a lower cost than any other highway in the system). The benefit of this highway is illustrated by the dramatically reduced cost incurred by the single traveler: by taking the short-cut, one traveler can traverse the network at a cost of 31 units ($2 V_1 + V_3$). Adding a new road has seemingly reduced the traversal cost dramatically.

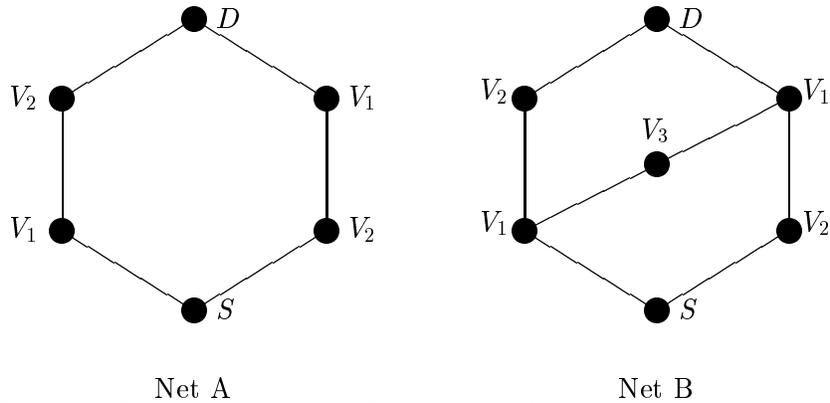

Net A        Net B

Figure 2: Hex network with $V_1 = 10x$ ; $V_2 = 50 + x$ ; $V_3 = 10 + x$

However consider what happens when six travelers are on the highways in net B. If each agent uses an ISPA, then at equilibrium each of the three possible paths contains two travelers.[3] Due to overlaps in the paths however, this results in each traveler incurring a cost of 92 units, which is higher than than what they incurred *before* the new highway was built. The net effect of adding a new road is to increase the cost incurred by *every* traveler. This phenomenon is known as Braess' paradox.

---

2. A detailed discussion and proof of the suboptimality of LB is shown in appendix A. Since LB is not used in current systems and is hard to imagine ever being used, our experiments do not consider it; it is discussed here for pedagogical reasons.

3. We have in mind here the Nash equilibrium for this problem, where no traveler (or equivalently, no router) can gain advantage by changing strategies.

## 3. Mathematics of Collective Intelligence

One common solution to these types of side-effect problems is to have particular agents of the network (e.g., a "network manager" Korilis, Lazar, & Orda, 1995) dictate certain choices to other agents. This solution can incur major brittleness and scaling problems however. Another kind of approach, which avoids the problems of a centralized manager, is to provide the agents with extra incentives that can induce them to take actions that are undesirable to them from a strict SPA sense. Such incentive can be in the form of "taxes" or "tolls" added to the costs associated with traversing particular links to discourage the use of those links. Such schemes in which tolls are superimposed on the agents' goals are a special case of the more general approach of replacing the goal of each agent with a new goal. These new goals are specifically tailored so that if they are collectively met the system maximizes throughput. *A priori*, a agent's goal need have no particular relation with the SPA-type cost incurred by that agent's packets. Intuitively, in this approach, we provide each agent with a goal that is "aligned" with the global objective, with no separate concern for of that goal's relation to the SPA-type cost incurred by the traffic routed by that agent.

In this section, we summarize the salient aspects of a Collective Intelligences (COIN) (Wolpert, Wheeler, & Tumer, 2000; Wolpert & Tumer, 1999). In this paper we consider systems that consist of a set of agents, connected in a network, evolving across a set of discrete, consecutive time steps, $t \in \{0, 1, ...\}$. Without loss of generality, we let all relevant characteristics of a agent $\eta$ at time $t$ — including its internal parameters at that time as well as its externally visible actions — be encapsulated by a Euclidean vector $\underline{\zeta}_{\eta,t}$ with components $\underline{\zeta}_{\eta,t;i}$. We call this the "state" of agent $\eta$ at time $t$, and let $\underline{\zeta}_{,t}$ be the state of all agents at time $t$, while $\underline{\zeta}$ is the state of all agent across all time.

**World utility**, $G(\underline{\zeta})$, is a function of the state of all agents across all time. When $\eta$ is an agent that uses a Machine Learning (ML) algorithm to "try to increase" its **private utility**, we write that private utility as $g_{\eta}(\zeta)$, or more generally, to allow that utility to vary in time, $g_{\eta,\tau}(\underline{\zeta})$.

We assume that $\underline{\zeta}$ encompasses all physically relevant variables, so that the dynamics of the system is deterministic (though of course imprecisely known to anyone trying to control the system). Note that this means that *all* characteristics of an agent $\eta$ at $t = 0$ that affects the ensuing dynamics of the system must be included in $\underline{\zeta}_{\eta,0}$. For ML-based agents, this includes in particular the algorithmic specification of its private utility, typically in the physical form of some computer code (the mathematics can be generalized beyond ML-based agents, as elaborated in Wolpert & Tumer, 1999).

Here we focus on the case where our goal, as COIN designers, is to maximize world utility through the proper selection of private utility functions. Intuitively, the idea is to choose private utilities that are aligned with the world utility, and that also have the property that it is relatively easy for us to configure each agent so that the associated private utility achieves a large value. In this paper, all utilities we consider are of the form $\sum_{t \geq \tau} R_t(\underline{\zeta}_{,t})$ for **reward functions** $R_t$ (simply $\sum_t R_t(\underline{\zeta}_{,t})$ for non-time-varying utilities). From now on, we will only consider world utilities whose associated set of $\{R_t\}$ are all time-translations of one another. In particular, as shown below, overall network throughput is expressible this way.

We need a formal definition of the concept of having private utilities be "aligned" with $G$. Constructing such a formalization is a subtle exercise. For example, consider systems where the world utility is the sum of the private utilities of the individual agents. This might seem a reasonable candidate for an example of "aligned" utilities. However such systems are examples of the more general class of systems that are "weakly trivial". It is well-known that in weakly trivial systems each individual agent greedily trying to maximize its own utility can lead to the tragedy of the commons (Hardin, 1968; Crowe, 1969) and actually *minimize* $G$. In particular, this can be the case when private utilities are independent of time and $G = \sum_\eta g_\eta$. Evidently, at a minimum, having $G = \sum_\eta g_\eta$ is not sufficient to ensure that we have "aligned" utilities; some alternative formalization of the concept is needed. Note that in the simple network discussed in Section 2.1, the utilities are weakly trivial, since $G(\vec{x}, \vec{y}) = g_X(\vec{x}) + g_y(\vec{y})$. This provides another perspective on the suboptimality of ISPA in that network.

A more careful alternative formalization of the notion of aligned utilities is the concept of "factored" systems. A system is **factored** at time $\tau$ when the following holds for each agent $\eta$ individually: A change at time $\tau$ to the state of $\eta$ alone, when propagated across time, will result in an increased value of $g_{\eta,\tau}(\underline{\zeta})$ if and only if it results in an increase for $G(\underline{\zeta})$ (Wolpert & Tumer, 1999).

For a factored system, the side-effects of any change to $\eta$'s $t = \tau$ state that increases its private utility cannot decrease world utility. There are no restrictions though on the effects of that change on the private utilities of other agents and/or times. In particular, we don't preclude an agent's algorithm at two different times from "working at cross-purposes" to each other, so long as at both moments the agent is working to improve $G$. In game-theoretic terms, in factored systems optimal global behavior corresponds to the agents' always being at a private utility Nash equilibrium (Fudenberg & Tirole, 1991). In this sense, there can be no tragedy of the commons for a factored system. As a trivial example, a system is factored for $g_{\eta,\tau} = G \, \forall \eta$, a system conventionally called a 'team game'.

Furthermore, if our system is factored with respect to private utilities $\{g_{\eta,\tau}\}$, we want each agent to be in a state at time $\tau$ that induces as high a value of the associated private utility as possible (given the initial states of the other agents). Assume $\eta$ is ML-based and able to achieve fairly large values of most private utilities we are likely to set it for time $\tau$, i.e., assume that given that private utility $g_{\eta,\tau}$, the rest of the components of $\underline{\zeta}_{\eta,\tau}$ are set by $\eta$'s algorithm in such a way so as to achieve a relatively high value of $g_{\eta,\tau}$. So our problem becomes determining for what $\{g_{\eta,\tau}\}$ the agents will best be able to achieve high $g_\eta$ (subject to each other's actions) while also causing dynamics that is factored for $G$ and the $\{g_{\eta,\tau}\}$.

Define the **effect set** of the agent-time pair $(\eta, \tau)$ at $\underline{\zeta}$, $C^{eff}_{(\eta,\tau)}(\underline{\zeta})$, as the set of all agents $\underline{\zeta}_{\eta',t}$ which under the forward dynamics of the system have non-zero partial derivative with respect to the state of agent $\eta$ at $t = \tau$. Intuitively, $(\eta, \tau)$'s effect set is the set of the states of all agents $\underline{\zeta}_{\eta',t \geq \tau}$ that would be affected by a change in the state of agent $\eta$ at time $\tau$.

Next, for any set $\sigma$ of agents $(\eta', t)$, define $\mathrm{CL}_\sigma(\underline{\zeta})$ as the "virtual" vector formed by clamping the components of the vector $\underline{\zeta}$ delineated in $\sigma$ to an arbitrary fixed value, which in this paper is set to 0. [4] This operation creates a new state vector (e.g., worldline) where the clamped components of that worldline (e.g., one player's action at a particular time step) are "zeroed" (e.g., removed from the system).

The value of the **wonderful life utility** (WLU for short) for $\sigma$ is defined as:

$$WLU_\sigma(\underline{\zeta}) \equiv G(\underline{\zeta}) - G(\mathrm{CL}_\sigma(\underline{\zeta})). \tag{1}$$

In particular, we are interested in the WLU for the effect set of agent-time pair $(\eta, \tau)$. This WLU is the difference between the actual world utility and the virtual world utility where all agent-time pairs that are affected by $(\eta, \tau)$ have been clamped to a zero state while the rest of $\underline{\zeta}$ is left unchanged.

Since we are clamping to $\vec{0}$, we can loosely view $(\eta, \tau)$'s effect set WLU as analogous to the change in world utility that would have arisen if $(\eta, \tau)$ "had never existed", hence the name of this utility - cf. the Frank Capra movie. Note however, that CL is a purely "fictional", counter-factual operator, in that it produces a new $\underline{\zeta}$ without taking into account the system's dynamics. The sequence of states the agent-time pairs in $\sigma$ are clamped to in constructing the WLU need not be consistent with the dynamical laws of the system. This dynamics-independence is a crucial strength of the WLU. It means that to evaluate the WLU we do *not* try to infer how the system would have evolved if agent $\eta$'s state were set to $\vec{0}$ at time $\tau$ and the system evolved from there. So long as we know $\underline{\zeta}$, extending over all time, $\sigma$, and the function $G$, we know the value of WLU.

As mentioned above, regardless of the system dynamics, having $g_{\eta,\tau} = G \; \forall \eta$ means the system is factored at time $\tau$.

**Theorem:** Regardless of the system dynamics, setting $g_{\eta,\tau} = WLU_{C^{eff}_{(\eta,\tau)}} \; \forall \eta$ results in a factored system at time $\tau$.

**Proof:** The second term, $G(\mathrm{CL}_{C^{eff}_{(\eta,\tau)}}(\underline{\zeta}))$ is, by definition, independent of $\underline{\zeta}_{\eta,\tau}$. Therefore a change to only the $(\eta, \tau)$ component of $\underline{\zeta}$ will only affect the first term, $G(\underline{\zeta})$. Therefore the effect of such a change on the value of the world utility is the same as its effect on the value of the wonderful life utility. **QED.**

Since factoredness does not distinguish the team game and wonderful life utilities, we need some other means of deciding which to use as our choice of $\{g_{\eta,\tau}\}$. To determine this, note that since each agent is operating in a large system, it may experience difficulty discerning the effects of its actions on $G$ when $G$ sensitively depends on all the agents in the system. Therefore each $\eta$ may have difficulty learning from past experience what to do to achieve high $g_{\eta,\tau}$ when $g_{\eta,\tau} = G$. In particular, in routing in large networks, having private rewards given by the world reward functions means that to provide each router with its reward at each time step we need to provide it the full throughput of the entire network at that step. This is usually infeasible in practice. Even if it weren't though, using these private utilities would mean that the routers face a very difficult task in trying to discern

---

the effect of their actions on their rewards, and therefore would likely be unable to learn their best routing strategies.

This problem can be mitigated by using effect set WLU as the private utility, since the subtraction of the clamped term removes much of the "noise" of the activity of other agents, leaving only the underlying "signal" of how the agent in question affects the utility (this reasoning is formalized as the concept of "learnability" in Wolpert & Tumer, 1999). Accordingly, one would expect that setting private utilities to WLU's ought to result in better performance than having $g_{\eta,\tau} = G \; \forall \eta, \tau$. This is the primary theoretical consideration that we leverage in the COIN techniques investigated in this paper.

In practice, we will sometimes only be able to estimate the "primary", most prominent portion of the effect set. Technically, the associated WLU is not the effect set WLU, and therefore not exactly factored. However assuming that that associated WLU is close enough to being factored, we would expect the advantage in learnability with such a WLU to still result in better performance than would using $g_{\eta,\tau} = G \; \forall \eta, \tau$ (see Wolpert et al., 2000; Wolpert & Tumer, 1999). Indeed, for the sake of improving learnability, sometimes we will elect to exclude certain agent-time pairs from our estimate of the effect set of $(\eta, \tau)$, even if we are sure that that are affected by $\underline{\varsigma}_{\eta,\tau}$. This will be the case if we expect that the changes in $G$ due to varying $\underline{\varsigma}_{\eta,\tau}$ that are "mediated" through those agent-time pairs are relatively insignificant, and therefore effectively constitute noise for the learning process, so that their effect on learnability is more important than their effect on factoredness.

## 4. Collective Intelligence for Network Routing

In this section, we use the theory summarized in Section 3 to derive individual goals for each router, in the form of private utility functions to be maximized by appropriate choice of routing decisions. The routers tried to achieve those maximizations by using algorithms that only require limited knowledge of the state of the network (in particular knowledge that is readily available to routers in common real data networks). In our simulations each router used a Memory Based (MB) machine learning algorithm (nearest neighbor) to make routing decisions. More precisely, for each potential routing decision, the routers look for the past state that most closely closely matches their current state (e.g., load). They then assign an "estimated" utility value to each potential routing decision and select the action with the highest estimated utility value. We call this algorithm an MB COIN[5].

### 4.1 Model Description

To apply the COIN formalism to a network routing model, we must formally describe that as a set of deterministically evolving vectors $\underline{\varsigma}_t$. In the model used in this paper, at any time step all traffic at a router is a set of pairs of integer-valued traffic amounts and associated ultimate destination tags. At each such time step $t$, each router $r$ sums the integer-valued components of its current traffic at that time step (one component for each

---

5. Relatively minor details of the algorithm concerning exploration/exploitation issues along with a "steering" parameter are discussed at the end of this section.

ultimate destination) to get its **instantaneous load**. We write that load as:

$$z_r(t) \equiv \sum_d x_{r,d}(t),$$

where the index $d$ runs over ultimate destinations, and $x_{r,d}(t)$ is the total traffic at time $t$ going from $r$ towards $d$. After its instantaneous load at time $t$ is evaluated, the router sends all its traffic to the next downstream routers, in a manner governed by the underlying routing algorithm. We indicate such "next routers" by writing:

$$x_{r,d}(t) = \sum_{r'} x_{r,d,r'}(t),$$

where $r'$ is the next router for traffic $(r, d)$, i.e., the first stop on the path to be followed from router $r$ to ultimate destination $d$. After all such routed traffic goes to those next downstream routers, the cycle repeats itself, until all traffic reaches its destinations.

In our simulations, for simplicity, traffic was only introduced into the system (at the **source routers**) at the beginning of successive disjoint **waves** of $L$ consecutive time steps each[6]. We use $\kappa(t)$ to indicate either the integer-valued wave number associated with time $t$ or the set of all times in that wave, as the context indicates.

In a real network, the cost of traversing a router depends on "after-effects" of recent instantaneous loads, as well as the current instantaneous load. To simulate this effect, we use time-averaged values of the load at a router rather than instantaneous load to determine the cost a packet incurs in traversing that router. More formally, we define the router's **windowed load**, $Z_r(t)$, as the running average of that router's load value over a window of the previous $W$ timesteps ($W$ is always set to an integer multiple of $L$):

$$Z_r(t) \equiv \frac{1}{W} \sum_{t'=t-W+1}^{t} z_r(t') = \sum_d X_{r,d}(t),$$

where the value of $X_{r,d}(t)$ is set by

$$X_{r,d}(t) = \frac{1}{W} \sum_{t'=t-W+1}^{t} x_{r,d}(t').$$

Intuitively, for large enough $W$, using such a window to determine costs across routers means that typically those costs will only change substantially over time scales significantly larger than that of the individual routing decisions. Formally, the windowed load is the argument to a **load-to-cost** function, $V(\cdot)$, which provides the **cost** accrued at time $t$ by each packet traversing the router at this timestep. That is, at time $t$, the cost for each packet to traverse router $r$ is given by $V(Z_r(t))$[7]. Note that in our model, the costs are accrued at the routers, not the links. Also note that for simplicity we do not physically instantiate the cost as a temporal delay in crossing a router. Different routers have different $V(\cdot)$, to reflect the fact that real networks have differences in router software and hardware (response time, queue length, processing speed etc). For simplicity, $W$ is the same for all routers however. With these definitions, world utility is given by

$$
\begin{aligned}
G(\underline{\zeta}) & = \sum_{t,r} z_r(t) \; V_r(Z_r(t)) \\
& = \sum_{t,r,d} x_{r,d}(t) V_r(Z_r(t)) \\
& = \sum_{t,r,d} x_{r,d}(t) V_r \left( \frac{1}{W} \sum_{t'=t-W+1}^{t} \sum_{d'} x_{r,d'}(t') \right) \\
& = \sum_{t,r,d} x_{r,d}(t) V_r \left( \sum_{d'} X_{r,d'}(t) \right) \; .
\end{aligned}
\tag{2}
$$

---

6. $L$ was always chosen to be the minimal number necessary for all traffic to reach its destination before the next wave of traffic is initiated.

7. We also introduce "dummy routers" denoted by $V_0(\cdot) = 0$ which help in translating the mathematics into the simulations. Omitting them will have no effect on the simulations.

Our equation for $G$ explicitly demonstrates that, as claimed above, in our representation we can express $G(\underline{\zeta})$ as a sum of rewards, $\sum_t R_t(\underline{\zeta}_{,t})$, where $R(\underline{\zeta}_{,t})$ can be written as function of a pair of $(r,d)$-indexed vectors:

$$
R_t(x_{r,d}(t), X_{r,d}(t)) = \sum_{r,d} x_{r,d}(t) V_r \left( \sum_{d'} X_{r,d'}(t) \right) .
$$

Also as claimed, the $R_t$ are temporal translations of one another.

Given this model, some of the components of $\underline{\zeta}_{,t}$ must be identified with the values $x_{r,d,r'}(t) \; \forall \, r, d, r'$ and $t$, since those $x$'s are set by the actions the agents will take. Since all arguments of $G$ must be components of $\underline{\zeta}$, we also include the $X_{r,d}(t) \; \forall r, d, t$ as components of $\underline{\zeta}_{,t}$. Formally, for routing based on $\overline{\text{ML}}$ agents, the internal parameters of the ML agents must also be included in $\underline{\zeta}$. This is because those parameters affect the routing, and in turn are affected by it. So to have $\underline{\zeta}$ evolve deterministically, since it includes the routing variables, it must also contain internal parameters of the agents. We won't have any need to explicitly delineate such variables here however, and will mostly phrase the discussion as though there were no such internal parameters.

Now the values $\{x_{r,d,r'}(t-1)\} \; \forall r, d, r'$ specify the values $\{x_{r,d}(t)\} \; \forall r, d$ directly. Therefore, in concert with the $\{x_{r,d}(t' < t)\}$, they also set the $\{X_{r,d}(t)\}$ directly. Moreover in our simulations the decisions $\{x_{r,d,r'}(t)\} \; \forall r, d, r'$ fixed by the routing algorithms at all times $t$ are given by a fixed function of the $\{x_{r,d}(t)\}$ and the $\{Z_r(t) = \sum_{d'} X_{r,d'}(t)\}$. So in point of fact we can map the set of $\{x_{r,d,r'}(t-1), X_{r,d'}(t)\} \; \forall r, d, r'$ to the full set $\{x_{r,d,r'}(t)\} \; \forall r, d, r'$, not just to $\{x_{r,d}(t)\}$. Accordingly, the $x_{r,d,r'}$ undergo deterministic evolution. Since their values across time set all the values of the $X_{r,d}(t)$ across time, we see that the entire set of the components of $\underline{\zeta}_{,t}$ undergo deterministic evolution in this representation, as required.

For evaluating the wonderful life utility we will need to group the components of $\underline{\zeta}_{,t}$ into disjoint agents $\eta$. Here we will have two types of agent, both types being indexed by router-destination pairs. For each such agent index $(r,d)$, the first agent type is the variable $X_{r,d}(t)$, and the second agent type is the Euclidean vector with components indexed by $r'$, $(x_{r,d})_{r'}(t)$. In setting "actions" we are concerned with setting the states of the agents of the second type. Accordingly, our learners will all be associated with agents of this second type. Unless explicitly indicated otherwise, from now on we will implicitly have that second type of agent in mind whenever we refer to a "agent" or use the symbol $\eta$.

## 4.2 ISPA Routing and COIN Routing

Based on the COIN formalism presented in Section 3 and the model described above, we now present the ISPA and COIN-based routing algorithms. At time step $t$, ISPA has access to all the windowed loads at time step $t-1$ (i.e., it has access to $Z_r(t-1) \; \forall r$), and assumes that those values will remain the same at all times $\geq t$. Note that for large window sizes and times close to $t$, this assumption is arbitrarily accurate. Using this assumption, in ISPA, each router sends packets along the path that it calculates will minimize the costs accumulated by its packets.

The COIN-based routing algorithms, in contrast, do not have such direct access to the $Z_r$. So to evaluate the WLU for a agent $(r, d)$ at any time $\tau$, such an algorithm must estimate the (primary members of the) associated effect set. This means determining what components of $\underline{\zeta}$ will, under the dynamics of the system, be changed by altering any of the components of the vector $x_{r,d}(\tau)$.

As a first approximation, we will ignore effects on traffic that changing $x_{r,d,r'}(\tau)$ may have that are "mediated" by the learning algorithms running in the system. That is, we ignore changes that arise due to the the effects that changing $x_{r,d,r'}(\tau)$ has on rewards, changes which induce changes in future training sets, which then in turn get mapped to changes in the $\{x_{r,d,r'}(t)\}$ (and therefore the $\{X_{r,d}(t)\}$) via the learning algorithms running on the agents.

As another approximation, we will ignore effects mediated by the routing algorithms' observations of the state of the network. That is, we ignore changes in the $\{x_{r'',d',r'''}(t)\}$ that varying $x_{r,d}(\tau)$ may cause due to associated changes in the state of the network perceived by $(r'', d')$'s routing algorithm, changes that in turn cause that algorithm to modify its routing decisions accordingly. We only consider the behavior of those routing algorithms that are (potentially) directly affected by $x_{r,d}(\tau)$ in that they (potentially) have to route packets that, at time $\tau$, passed through $r$ on the way to $d$. So in particular we ignore effects of $x_{r,d}(\tau)$ on the $\{x_{r'',d' \neq d,r'''}(t)\}$.

Since all packets routed in a wave arrive at their destinations by the end of the wave, these approximations mean that the only $x_{r'',d'',r'''}(t)$ that are in our estimate for $x_{r,d}(\tau)$'s effect set have $t$ in the same wave as $\tau$. These are the only ones that are, potentially, directly affected by the $\{x_{r,d,r'}(t)\}$ by "chaining together" the sequence of $x_{r'',d'',r'''}(t)$ that get the packets in $x_{r,d}(t)$ to their ultimate destination. Due to the wave nature of our simulations though, the only $x_{r'',d'',r'''}(t)$ within $\tau$'s wave that are affected by $x_{r,d}(\tau)$ all have $d'' = d$. For reasons of coding simplicity, we do not concern ourselves with whether $t < \tau$ within a given wave and then exclude some $x_{r'',d'',r'''}(t)$ accordingly. In other words, all $t$ within $\tau$'s wave are treated equally.

So one set of members of $x_{r,d}(\tau)$'s effect set is $\{x_{r'',d,r'''}(t) \; \forall r'', d, r''', t \in \kappa(\tau)\}$. Note that some of these members will be relatively unaffected by $x_{r,d}(\tau)$ (e.g., those with $r''$ far in the net away from $r$). Again for simplicity, we do not try to determine these and exclude them. As with keeping the $x_{r'',d,r'''}(t < \tau)$, this inclusion of extra agents in our estimate of the effect set should hurt learnability, but in general should not hurt factoredness. Therefore it should delay how quickly the learners determine their optimal policies, but it won't affect the quality (for $G$) of those policies finally arrived at. Note also that trying to determine whether some particular $x_{r'',d,r'''}(t \in \kappa(\tau))$ should be included in $x_{r,d}(\tau)$'s effect set would mean, in part, determining whether packets routed from $(r,d)$ would have reached $r''$ if $(r,d)$ had made some routing decision different from the one it actually made. This would be a non-trivial exercise, in general.

In contrast to the case with the $x_{r'',d',r'''}(t)$, there are $X_{r'',d'}(t)$ with $t$ in the future of $\tau$'s wave that both are affected by $x_{r,d}(t)$ and also are not excluded by any of our approximations so far. In particular, the $X_{r'',d}(t)$ with either $r'' = r$ or $r''$ one hop away from $r$ will be directly affected by $x_{r,d}(t)$, for $t \in \cup_{i=0}^{W-1} \kappa(\tau + iL)$ (cf. the definition of the $X$ variables). For simplicity, we restrict consideration of such $X_{r'',d}$ variables to those with the same router as $r$, $r'' = r$.

This final estimate for the effect set is clearly rather poor — presumably results better than those presented below would accrue to use of a more accurate effect set. However it's worth bearing in mind that there is a "self-stabilizing" nature to the choice of effect sets, when used in conjunction with effect set WLU's. This nature is mediated by the learning algorithms. If one assigns the same utility function to two agents, then the reward one agent gets will be determined in part by what the other one does. So as it modifies its behavior to try to increase its reward, that first agent will be modifying its behavior in a way dependent on what the other agent does. In other words, if two agents are given the same WLU because they are estimated to be in each other's effect set, then *ipso facto* they will be in each other's effect set.

Using our estimate for the effect set, the WLU for $(\eta, \tau)$ is given by the difference between the total cost accrued in $\tau$'s wave by all agents in the network and the cost accrued by agents when all agents sharing $\eta$'s destination are "erased." More precisely, any agent $\eta$ that has a destination $d$ will have the following effect set WLU's, $g_{\eta,\tau}$:

$$
\begin{aligned}
g_{\eta,\tau}(\underline{\zeta}) =\ & G(\underline{\zeta}) - G(\mathrm{CL}_{C^{eff}_{(\eta,\tau)}}(\underline{\zeta})) \\
=\ & \sum_{t,r',d'} x_{r',d'}(t)\, V_{r'}\left( \sum_{d'} X_{r',d'}(t) \right) - \sum_{t,r',d'} \left[ x_{r',d'}(t)(1 - I(t \in \kappa(\tau)) I(d' = d)) \right] \\
& \times\ V_{r'}\left( \sum_{d''} \left[\, X_{r',d''}(t)\ \ (1 - I(t \in \cup_{i=0}^{W-1} \kappa(\tau + iL)) I(d'' = d)) \,\right] \right) \\
=\ & \sum_{t \in \kappa(\tau)} \sum_{r'} \left( \sum_{d'} x_{r',d'}(t)\ V_{r'}(\sum_{d''} X_{r',d''}(t)) - \sum_{d' \neq d} x_{r',d'}(t)\, V_{r'}(\sum_{d'' \neq d} X_{r',d''}(t)) \right) \\
& + \sum_{t \in \cup_{i=1}^{W-1} \kappa(\tau + iL)} \sum_{r'} \left( \sum_{d'} x_{r',d'}(t)\, [V_{r'}(\sum_{d''} X_{r',d''}(t)) - V_{r'}(\sum_{d'' \neq d} X_{r',d''}(t))] \right) \quad (3)
\end{aligned}
$$

where $I(.)$ is the indicator function that equals 1 if its argument is true, 0 otherwise.

To allow the learner to receive feedback concerning its actions in a wave immediately following that wave rather than wait for $\sim WL$ time steps, we will approximate the second sum in that last equality, the one over times following $\tau$'s wave, as zero. There is another way we can view the resultant expression, rather than as an approximation to the effect set WLU. That is to view it as the exact WLU of an approximation to the effect set, an approximation which ignores effects on future windowed loads of clamping a current traffic level. Regardless of what view we adopt, presumably better performance could be achieved if we did not implement this approximation.

Given this approximation, our WLU becomes a wave-indexed time-translation-invariant WL "reward function" (WLR):

$$g_{\eta,\tau}(\underline{\zeta}_{,t\in\kappa(\tau)}) = \sum_{t\in\kappa(\tau),r'} \left( \sum_{d'} x_{r',d'}(t) \ V_{r'}(\sum_{d''} X_{r',d''}(t)) \right.$$
$$\left. - \sum_{d'\neq d} x_{r',d'}(t) \ V_{r'}(\sum_{d''\neq d} X_{r',d''}(t)) \right). \tag{4}$$

Notice that traffic going from a router $r' \neq r$ to a destination $d' \neq d$ affects the value of the WLR for agent $(r,d)$. This reflects the fact that WLR takes into account side-effects of $(r,d)$'s actions on other agents. Note also that each $r'$-indexed term contributing to the WLR can be computed by the associated router $r'$ separately, from information available to that router. Subsequently those terms can be propagated through the network to $\eta$, in much the same way as routing tables updates are propagated.

Given this choice of private utility, we must next specify how the COIN-based routing algorithm collects the initial data that (in conjunction with this utility) is to be used to guide the initial routing decisions that every agent with more than one routing option must make. In our experiments that data was collected during a preliminary running of an ISPA. In this preliminary stage, the routing decisions are made using the ISPA, but the resulting actions are "scored" using the WLR given by Equation 3. We use the ISPA to generate the routing decisions in the initial data since it is likely in practice that some kind of SPA will be the routing algorithm running prior to "turning on" the COIN algorithm. Alternately one can generate the initial data's routing decisions by having the routers make random decisions, or by having them implement a sequence of decisions that "sweeps" across a grid through the possible set of actions. The data collected in this stage provides us with initial input-output training sets to be used by the machine learning algorithm on each agent: for each router-destination agent, inputs are identified with windowed loads on outgoing links, and the associated WLR values for the destination in question are the outputs.

After sufficient initial data is collected using the ISPA, the system switches to using the COIN algorithm to make subsequent routing decisions. In this stage, each agent routes packets along the link that it estimates (based on the training set) would provide the best WLR. To perform the estimation, the MB COIN makes use of a single-nearest-neighbor algorithm as its learner. This algorithm simply guesses that the output that would ensue from any candidate input is the same as the output of the element of the training set that is the nearest neighbor (in input space) of that candidate input.[8] In other words, the learner finds the training set input-output pair whose input value (loads on outgoing links) is closest to that which would result from each potential routing decision. Then the learner assigns the WLR associated with that training data pair as the estimate for what WLR would result from said routing decision. These WLR values are then used to choose among those potential routing decisions. The input-output data generated under this algorithm is adding to the training set as it is generated.

---

8. This is a very simple learning algorithm, and we use it here only to demonstrate the potential practical feasibility of a COIN-based routing algorithm. The performance can presumably be improved if more sophisticated learning algorithms (e.g., Q-learning Sutton & Barto, 1998; Watkins & Dayan, 1992) are used.

In this routing algorithm, the routers only estimate how their routing decisions (as reflected in their loads at individual time steps) will affect their WLR values (based on many agents' loads). It is also possible to calculate *exactly* how the routing decisions affect the routers' WLR's if, unlike the MB COIN, we had full knowledge of the loads of all agents in the system. In a way similar to ISPA, for each router we can evaluate the exact WLR value that would ensue from each of its candidate actions, under the assumption that windowed loads on all other routers are the same one wave into the future as they are now. We call this algorithm for directly maximizing WLR (an algorithm we call the full knowledge COIN, or FK COIN).

Note that under the assumption behind the FK COIN, the action $\eta$ chooses in wave $\kappa(\tau)$ that maximizes WLR will also maximize the world reward. In other words, WL reward is perfectly factored with respect to (wave-indexed) world reward, even though the associated utilities are not related that way (due to inaccuracy in our estimate of the effect set). Due to this factoredness, the FK COIN is equivalent to load balancing on world rewards. Since LB in general results in inferior performance compared to LB over time, and since the FK COIN is equivalent to LB, one might expect that its performance is suboptimal. Intuitively, this suboptimality reflects the fact that one should not choose the action only with regard to its effect on current reward, but also with concern for the reward of future waves. In the language of the COIN framework, this suboptimality can be viewed as a restatement of the fact that for our inexactly estimated effect set, the system will not be perfectly factored.

The learning algorithm of the MB COIN as described is extraordinarily crude. In addition, the associated scheme for choosing an action is purely exploitative, with no exploration whatsoever. Rather than choose some particular more sophisticated scheme and tune it to fit our simulations, we emulated using more sophisticated algorithms *in general*. We did this by modifying the MB COIN algorithm to occasionally have the FK COIN determine a router's action rather than the purely greedy learner outlined above. The **steering parameter** discussed in Section 5.5 determines how often the routing decision is based on the MB COIN as opposed to the FK COIN.

## 5. Simulation Results

In practice, it is very difficult to implement either FK COIN or LB. In this section we use experiments to investigate behavior of algorithms that can conceivably be used in practice. More precisely, based on the model and routing algorithms discussed above, we have performed simulations to compare the performance of ISPA and MB COIN across a variety of networks, varying in size from five to eighteen routers. In all cases traffic was inserted into the network in a regular, non-stochastic manner at the sources. The results we report are averaged over 20 runs. We do not report error bars as they are all lower than 0.05.

In Sections 5.1 - 5.4 we analyze traffic patterns over four networks where ISPA suffers from the Braess' paradox. In contrast, the MB COIN almost never falls prey to the paradox for those networks (or for no networks we have investigated is the MB COIN significantly susceptible to Braess' paradox). Then in Section 5.5 we discuss the effect on the MB COIN's performance of the "steering" parameter which determines the intelligence of the MB COIN.[9]

## 5.1 Bootes Network

The first network type we investigate is shown in Figure 3. It is in many senses a trivial network, as in Net A, the sources do not even have any choices to make. The loads introduced at the sources do not change in time and are listed in Tables 1 and 2, along with the performances of our algorithms.

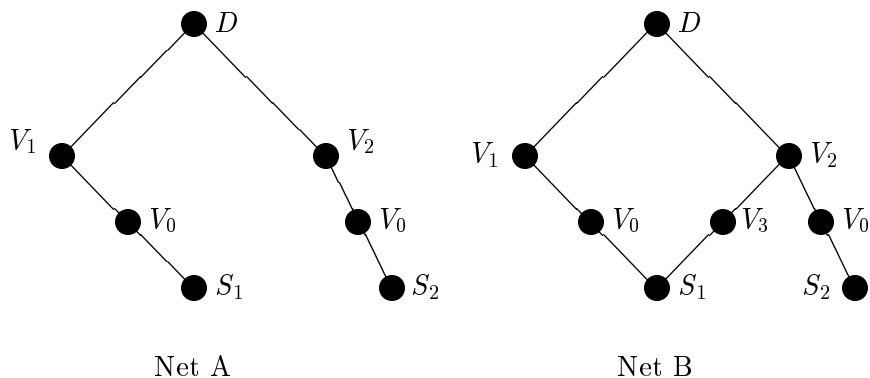

Net A                    Net B

Figure 3: Bootes Network

| Loads at $(S_1, S_2)$ | Net | ISPA | MB COIN |
|---|---|---|---|
| 1,1 | A | 6.35 | 6.35 |
|  | B | 8.35 | 5.93 |
| 2,1 | A | 8.07 | 8.07 |
|  | B | 10.40 | 7.88 |
| 2,2 | A | 9.55 | 9.55 |
|  | B | 10.88 | 9.71 |
| 4,2 | A | 10.41 | 10.41 |
|  | B | 11.55 | 10.41 |

Table 1: Average Per Packet Cost for BOOTES2 networks for $V_1 = 10 + log(1 + x)$ ; $V_2 = 4x^2$ ; $V_3 = log(1 + x)$ .

The MB COIN results are identical to the ISPA results in the absence of the additional link (Network A). However, Braess' paradox arises with ISPA, in that the addition of the new link in network B degrades the performance of the ISPA in six of the eight traffic regimes and load-to-cost functions investigated. The MB COIN on the other hand is only

9. In Sections 5.1 - 5.4, the steering parameter is set at 0.5.

| Loads at $(S_1, S_2)$ | Net | ISPA | MB COIN |
|:---:|:---:|:---:|:---:|
| 1,1 | A | 30.35 | 30.35 |
|  | B | 20.35 | 20.35 |
| 2,2 | A | 35.55 | 35.55 |
|  | B | 40.55 | 34.99 |
| 4,2 | A | 41.07 | 41.07 |
|  | B | 50.47 | 44.13 |
| 6,3 | A | 44.63 | 44.63 |
|  | B | 51.40 | 44.63 |

Table 2: Average Per Packet Cost for BOOTES4 network for $V_1 = 50 + log(1 + x)$ ; $V_2 = 10x$ ; $V_3 = log(1 + x)$ .

hurt by the addition of the new link once, and manages to gainfully exploit it seven times. When their behavior is analyzed infinitesimally, the MB COIN either uses the additional link efficiently or chooses to ignore it in those seven cases. Moreover, the MB COIN's performance with the additional link is always better than the ISPA's. For example, adding the new link causes a degradation of the performance by as much as 30 % (loads = $\{2, 1\}$) for the ISPA, whereas for the same load vector MB COIN performance improves by 7 %.

## 5.2 Hex Network

In this section we revisit the network first discussed in Section 2.1 (redrawn in Figure 4 to include the dummy agents). In Table 3 we give full results for the load-to-delay functions discussed in that section. We then use load-to-cost functions which are qualitatively similar to those discussed in Section 2.1, but which incorporate non-linearities that better represent real router characteristics. That load-to-cost function and associated results are reported in Table 4.

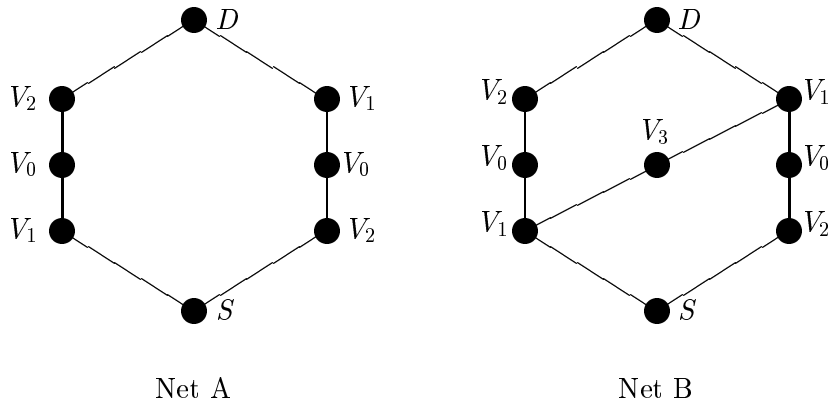

Figure 4: Hex network

This network demonstrates that while the addition of a new link may be beneficial in low traffic cases, it leads to bottlenecks in higher traffic regimes. For ISPA although the per packet cost for loads of 1 and 2 drop drastically when the new link is added, the per packet cost increases for higher loads. The MB COIN on the other hand uses the new link efficiently. Notice that the MB COIN's performance is slightly worse than that of the ISPA in the absence of the additional link. This is caused by the MB COIN having to use a learner to estimate the WLU values for potential actions whereas the ISPA simply has direct access to all the information it needs (costs at each link).

| Load | Net | ISPA | MB COIN |
|------|-----|-------|---------|
| 1 | A | 55.50 | 55.56 |
|   | B | 31.00 | 31.00 |
| 2 | A | 61.00 | 61.10 |
|   | B | 52.00 | 51.69 |
| 3 | A | 66.50 | 66.65 |
|   | B | 73.00 | 64.45 |
| 4 | A | 72.00 | 72.25 |
|   | B | 87.37 | 73.41 |

Table 3: Average Per Packet Cost for HEX network for $V_1 = 50 + x$ ; $V_2 = 10x$ ; $V_3 = 10 + x$ .

| Load | Net | ISPA | MB COIN |
|------|-----|-------|---------|
| 1 | A | 55.41 | 55.44 |
|   | B | 20.69 | 20.69 |
| 2 | A | 60.69 | 60.80 |
|   | B | 41.10 | 41.10 |
| 3 | A | 65.92 | 66.10 |
|   | B | 61.39 | 59.19 |
| 4 | A | 71.10 | 71.41 |
|   | B | 81.61 | 69.88 |

Table 4: Average Per Packet Cost for HEX network for $V_1 = 50 + log(1 + x)$ ; $V_2 = 10x$ ; $V_3 = log(1 + x)$ .

## 5.3 Butterfly Network

The next network we investigate is shown in Figure 5. It is an extension to the simple network discussed in Section 5.1. We now have doubled the size of the network and have three sources that have to route their packets to two destinations (packets originating at $S_1$ go to $D_1$, and packets originating at $S_2$ or $S_3$ go to $D_2$). Initially the two halves of the network have minimal contact, but with the addition of the extra link two sources from the two two halves of the network share a common router on their potential shortest path.

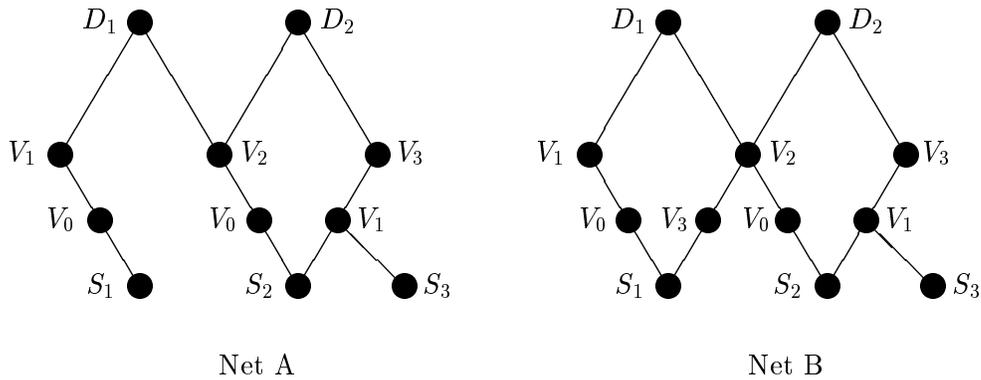

Figure 5: Butterfly Network

Table 5 presents two sets of results: first we present results for uniform traffic through all three sources, and then results for asymmetric traffic. For the first case, the Braess' paradox is apparent in the ISPA: adding the new link is beneficial for the network at low load levels where the average per packet cost is reduced by nearly 20%, but deleterious at higher levels. The MB COIN, on the other hand, provides the benefits of the added link for the low traffic levels, without suffering from deleterious effects at higher load levels.

| Loads $(S_1, S_2, S_3)$ | Net | ISPA | MB COIN |
|---|---|---|---|
| 1,1,1 | A | 112.1 | 112.7 |
| | B | 92.1 | 92.3 |
| 2,2,2 | A | 123.3 | 124.0 |
| | B | 133.3 | 122.5 |
| 4,4,4 | A | 144.8 | 142.6 |
| | B | 156.5 | 142.3 |
| 3,2,1 | A | 81.8 | 82.5 |
| | B | 99.5 | 81.0 |
| 6,4,2 | A | 96.0 | 94.1 |
| | B | 105.3 | 94.0 |
| 9,6,3 | A | 105.5 | 98.2 |
| | B | 106.7 | 98.8 |

Table 5: Average Per Packet Cost for BUTTERFLY network for $V_1 = 50 + log(1+x)$ ; $V_2 = 10x$ ; $V_3 = log(1 + x)$.

For the asymmetric traffic patterns, the added link causes a drop in performance for the ISPA, especially for low overall traffic levels. This is not true for the MB COIN. Notice also that in the high, asymmetric traffic regime, the ISPA performs significantly worse than the MB COIN even without the added link, showing that a bottleneck occurs on the right side of network alone (similar to the Braess' paradox observed in Section 5.1).

## 5.4 Ray Network

In all the networks and traffic regimes discussed so far the sources are the only routers with more than one routing option. The final network we investigate is a larger network where the number of routers with multiply options is significantly higher than in the previous networks. Figure 6 shows the initial network (Net A) and the "augmented" network (Net B), where new links have been added. The original network has relatively few choices for the routers, as packets are directed toward their destinations along "conduits." The new links are added in the augmented networks to provide new choices (crossing patterns) that could be beneficial if certain of the original conduits experience large costs.

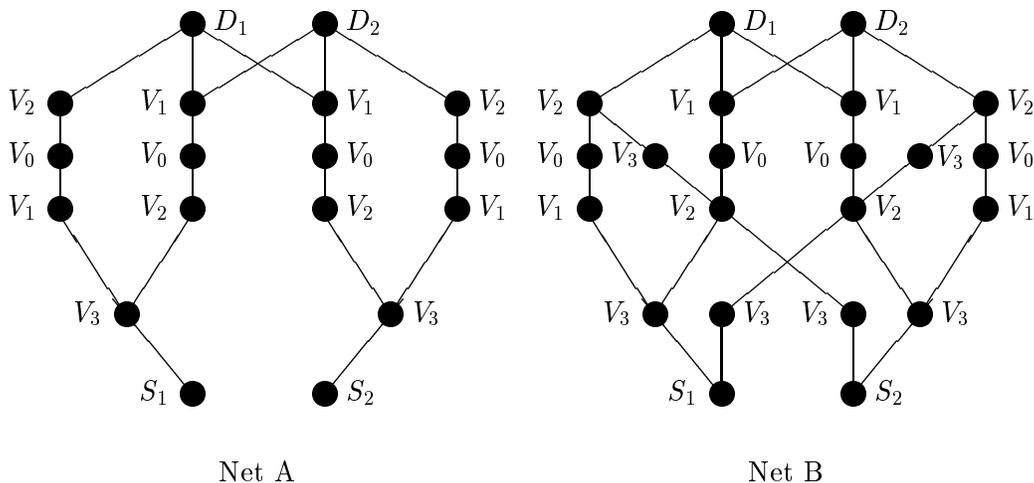

Net A                          Net B

Figure 6: Ray network

Table 6 shows the simulation results for these networks ($S_1$ and $S_2$ send packets to $D_1$ and $D_2$ respectively). At low load levels both the ISPA and the MB COIN use the new links effectively, although the MB COIN performs slightly worse. This is mainly caused by the difficulty encountered by the simple learner (single nearest neighbor algorithm) in quickly learning the traffic patterns in this large network. Unlike the ISPA however, the MB COIN avoids the Braess' paradox in all cases except the very high traffic regime. Moreover, even there, the effect is significantly milder than that encountered by the ISPA.

## 5.5 Steering the MB COIN

The final aspect of COIN-based routing we investigate is the impact of the choice for the value of the steering parameter. This parameter both controls the amount of exploration the algorithm performs and determines the "intelligence" of the MB COIN at estimating the surface directly calculated by the FK COIN. In Figures 7 - 8, the FK COIN results correspond to setting the steering parameter of the MB COIN to 1.0. This provides an upper bound on the performance that can be achieved though MB COIN.

For the HEX network (Figure 7), the performance at the worst setting for the MB COIN, which corresponds to no steering, is comparable to ISPA. In contrast, with moderate steering (0.5) the results are similar to that of the FK COIN, as the learner has more information to work with (arising from the extra parts of the input space represented in the training set due to the occasional use of the FK COIN), it bridges the gap between a suboptimal algorithm susceptible to Braess' paradox and one which efficiently avoids that paradox.

| Loads at $S_1 and S_2$) | Net | ISPA | MB COIN |
|---|---|---|---|
| 2,2 | A | 143.6 | 143.7 |
|  | B | 124.4 | 126.9 |
| 3,3 | A | 154.6 | 154.9 |
|  | B | 165.5 | 151.0 |
| 4,4 | A | 165.4 | 166.0 |
|  | B | 197.7 | 165.6 |
| 6,6 | A | 186.7 | 187.4 |
|  | B | 205.1 | 191.6 |

Table 6: Average Per Packet Cost for RAY network for $V_1 = 50 + log(1 + x)$ ; $V_2 = 10x$ ; $V_3 = 10 + log(1 + x)$.

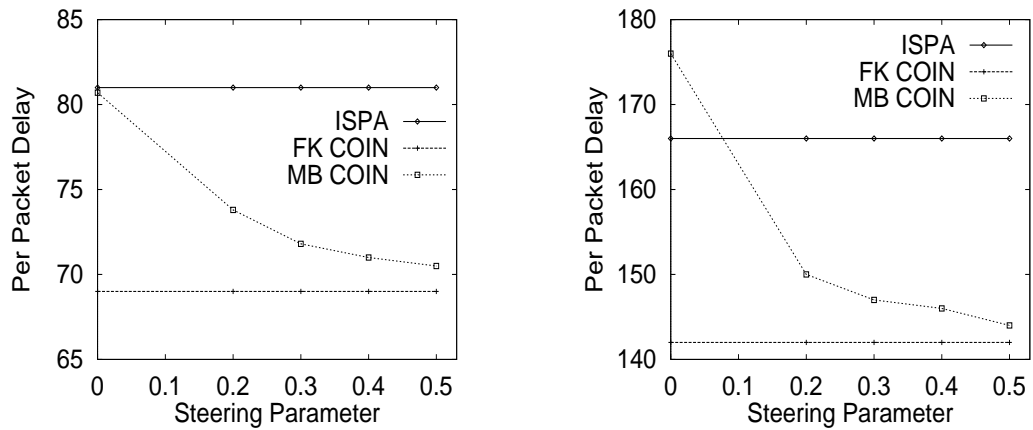

Figure 7: Impact of steering on Hex4 (left) and Ray4 (right) networks.

For the RAY network (Figure 7), the value of the steering parameter is more critical. With no steering at all, the MB COIN performs poorly in this network — even worse than ISPA. This is not surprising in that because there are many routing choices that affect the performance, the simple memory-based learner needs proper "seeding" to be able to perform well. Even with minimal steering though, the MB COIN quickly outperforms the ISPA.

Finally, for both the Butterfly and Bootes networks (Figure 8) the MB COIN needs very little steering to perform well. Although for the Butterfly network the performance of MB COIN improves slightly with more information, it is significantly better than the ISPA across the board.

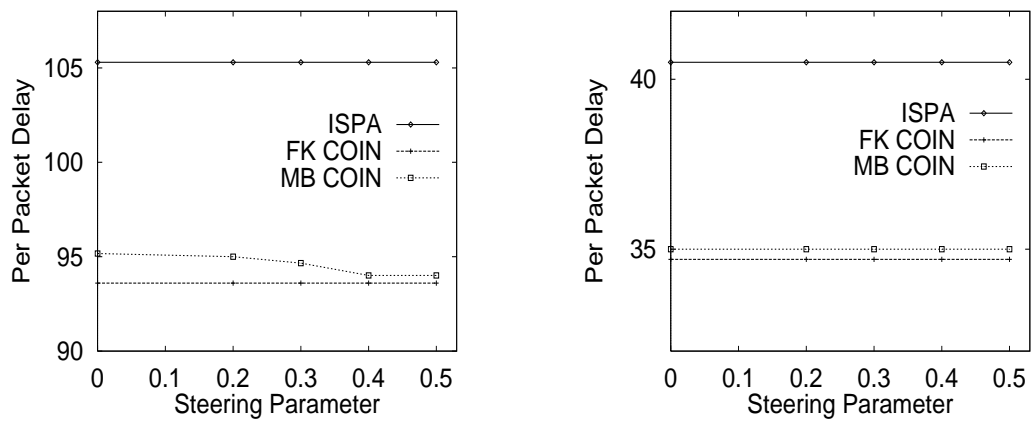

Figure 8: Impact of steering on Butterfly4 (left) and Bootes4 (right) networks.

## 6. Conclusion

Effective routing in a network is a fundamental problem in many fields, including data communications and transportation. Using a shortest path algorithm (SPA) on each of the routers to determine that router's decisions is a popular approach to this problem. However under certain circumstances it suffers from a number of undesirable effects. One such effect is Braess' paradox, where for the same pattern of introduced traffic into a network, increasing the capacity of that network results in *lower* overall throughput, due to the harmful side-effects of the decisions made by each router on the traffic in the rest of the system. Even the theoretical load-balancing algorithm, which addresses some of these effects to produce decisions that are optimal for any single moment of time, can still suffer from side-effects that result in sub-optimal performance. This is because such effects extend across time (i.e., what you do now affects performance later) as well as space.

The Collective Intelligence approach is a novel way of controlling distributed systems so as to avoid deleterious side-effects of routing decisions. The central idea is to have learning algorithms control the autonomous agents that constitute the overall distributed system. In such a Collective Intelligence (COIN), the central issue is to determine the personal objectives to be assigned to each of those autonomous agents. One wants to choose those goals so that the greedy pursuit of those goals by the associated learning algorithms leads to desirable behavior of the overall system. In this paper we have summarized the mathematics of designing such goals and derived a routing algorithm based on that mathematics.

We ran computer simulations to compare a COIN-based algorithm with an ideal SPA (whose performance upper-bounds all real-world SPA's) for routing. The COIN-based algorithm was severely handicapped. The estimation of the "effect sets" used by that algorithm was exceedingly crude. In addition, the learning algorithms of the agents were particularly unsophisticated, and therefore were not able to effectively maximize their individual performances. In contrast, the ideal SPA had access to more information concerning the state of the system than the (real-world-implementable) COIN did, information that no real-world SPA could access.

Despite these biases in favor of the ideal SPA, in our experiments the ideal SPA induced average costs as much as 32 % higher than the COIN-based algorithm. Furthermore the COIN-based algorithm almost always avoided the Braess' paradox that seriously diminished the performance of the SPA.

These techniques have also been very successfully employed in many other, non-routing domains, such as coordination of autonomous rovers (Tumer, Agogino, & Wolpert, 2002), combinatorial optimization, "congestion games" (Wolpert & Tumer, 2001), and control of data-upload from a planet (Wolpert, Sill, & Tumer, 2001). We conclude from these results that the techniques of the field of collective intelligence can be highly effective in designing the utility functions of the members of a MAS to ensure they work in a coordinated and efficient manner to optimize overall performance. We are currently investigating extensions of our COIN algorithm that involve novel goals for the agents, goals that are more "learnable" for the learning algorithms. We are also expanding the simulations to larger networks using a commercial event driven simulator. Future work will focus on not making the approximation that current traffic levels do not affect future windowed loads (Equation 3). It will also involve investigating better estimates of effect sets, in particular not including all agents with the same destination in one's effect set, and more generally using a more "fine-grained" representation of the agents, for example including each packet's originating source, to allow a more fine-grained effect set (and resultant WLU).

## Acknowledgments

The authors thank Joe Sill and the reviewers for their helpful comments.

## Appendix A. Suboptimality of Load-Balancing

In this appendix we we present an existence proof of the suboptimality of Load-Balancing (LB) by explicitly constructing a situation where conventional LB is suboptimal.

Consider a system with discrete time, in which the source agent $X$ under consideration must route one packet to the (fixed) destination at each time step. Presume further that no traffic from any source agent other than $X$ enters any of the agents $X$ sends to, so that traffic coming from $X$ is the sole source of any costs associated with $X$'s outbound links. Let $S(t)$ be the number of times our agent sent a packet down some link $A$ in the $W$ time steps preceding $t$, and take $s(t) = A, B$ to mean that the router uses link $A$ or $B$, respectively, at time $t$. Model queue backups and the like by having the cost to send a packet down link $A$ at time $t$ be $C_A(S(t)/W)$, and have the cost for our router to instead send the packet down link $B$ be $C_B(1 - S(t)/W)$, For simplicity we assume that both $C_A(.)$ and $C_B(.)$ are monotonically increasing functions of their arguments.

Restrict attention to agents that work by having $s(t) = A$ iff $S(t) \leq k$ for some real-valued threshold $k$. The LB algorithm will choose $s(t) = A$ iff $C_A(S(t)/W) \leq C_B(1 - S(t)/W)$. So the LB algorithm's behavior is indistinguishable from this kind of threshold algorithm, with $k$ set so that $C_A(k/W) = C_B(1 - k/W)$. (We implicitly assume that $C_A(.)$ and $C_B(.)$ are chosen so that such a solution exists for $1 < k < W - 1$.) The question is what $k$ should be to optimize total averaged cost across time, and in particular if that $k$ is the same as $k_{LB}$, the $k$ that LB uses.

Now as we go from one time step to the next, the routing decision made $W$ time steps ago drops out of the computation of $S(t)$, while the routing decision just made is newly included. In general, $S(t+1) = S(t) + 1$ if the router just used $A$ at time $t$ and used link $B$ at the time $W$ time steps into the past. On the other hand, $S(t+1) = S(t) - 1$ if the router just used $B$ and used $A$ $W$ time steps ago, while $S(t+1) = S(t)$ if the routing decision just made is the same as the routing decision $W$ time steps ago. So in general, $S(t)$ can only change by -1, 0, or +1 as we go from one time step to the next.

Consider cases where $1 < k < W - 1$, so that eventually the router must choose an $A$, and at some subsequent time $t^*$ the router switches from $A$ to $B$. At that time $s(t^*-1) = A$ and $s(t^*) = B$. This implies that $S(t^*-1) \leq k, S(t^*) > k$. Define the value $S(t^*-1)$ as $k^*$. Note that $S(t^*) = k^* + 1$, and $k - 1 < k^* \leq k$.

Now for any time $t'$, if $S(t') = k^* + 1$, $s(t'+1) = B$, and the only possible next values are $S(t'+1) = k^*$ or $S(t'+1) = k^* + 1$, depending on the old decision $s(t-W)$ that gets dropped out of the window. Similarly, if $S(t') = k^*$, $s(t'+1) = A$, and the only possible next values are $S(t'+1) = k^*$ or $S(t'+1) = k^* + 1$, again depending on the old decision being dropped. So we see that once $S(t') \in \{k^*, k^* + 1\}$, it stays there forever.

This means that because of the relationship between $k$ and $k^*$, in any interval of $W$ consecutive time steps subsequent to $t^*$, the number of packets sent along $A$ by router $X$ must be $\in (k-1, k+1]$. (Note that it is possible to send $k+1$ packets along $A$, but not $k-1$ packets. Therefore the number sent along $B$ must be $\in [W - (k+1), W - (k-1))$). Each time that a packet is sent along $A$ the cost incurred is the cost of link $A$ with average traffic level $S(t)/W$, $C_A(S(t)/W)$. Similarly, each time the link $B$ is chosen, the cost incurred is $C_B(1 - S(t)/W)$. Since $S(t) \in \{k^*, k^* + 1\}$, and both $C_A(.)$ and $C_B(.)$ are monotonically increasing, the cost for sending the packet down link $A \in (C_A((k-1)/W), C_A((k+1)/W]$, and that for sending it down link $B$ is contained in $[C_B(1-(k+1)/W), C_B(1-(k-1)/W))$.

Now we know that the choice of $A$ must have average frequency (across all time) between $k^*/W$ and $(k^*+1)/W$. Similarly, $B$ will have average frequency between $(1 - (k^*+1)/W)$ and $1 - k^*/W$. Accordingly, the average cost is bounded above by

$$\frac{k^*+1}{W} C_A \left( \frac{k+1}{W} \right) \ + \ \left( 1 - \frac{k^*}{W} \right) C_B \left( 1 - \frac{k-1}{W} \right) \ , \tag{5}$$

where the first term provides the maximum possible average cost for using link $A$, while the second term independently provides the maximum possible average cost for using link $B$. Note that the actual cost will be lower since the two frequencies in this bound, one for $A$ and one for $B$, cannot both have the values indicated. Because $k - 1 < k^* \leq k$ and since $1 - \frac{k-1}{W} = 1 + \frac{2}{W} - \frac{k+1}{W}$, our upper bound is itself bounded above by

$$\frac{k+1}{W} C_A \left( \frac{k+1}{W} \right) \ + \ \left( 1 + \frac{2}{W} - \frac{k+1}{W} \right) C_B \left( 1 + \frac{2}{W} - \frac{k+1}{W} \right) \ . \tag{6}$$

The optimal $k$ will result in an average cost lower than the minimum over all $k$ of the upper bound on average cost, given in Equation 6. So the average cost for the optimal $k$ is bounded above by the minimum over $k$ of this upper bound. Lable this argmin of Equation 6 $k'$.

Since other values of $k$ besides $k_{LB}$ result in behavior equivalent to LB, it does not suffice to simply test if $k' = k_{LB}$. Instead let us evaluate some lower bounds in a similar fashion to how we evaluated upper bounds. Using the average frequencies discussed above, the average cost is bounded below by:

$$\frac{k^*}{W} C_A \left( \frac{k-1}{W} \right) + \left( 1 - \frac{1}{W} - \frac{k^*}{W} \right) C_B \left( 1 - \frac{k+1}{W} \right) , \tag{7}$$

where the first term provides the minimum possible average cost for using link $A$, while the second term provides the minimum possible average cost for using link $B$. Again, because $k - 1 < k^* \leq k$, the term is Equation 7 is further bounded below by

$$\frac{k-1}{W} C_A \left( \frac{k-1}{W} \right) + \left( 1 - \frac{2}{W} - \frac{k-1}{W} \right) C_B \left( 1 - \frac{2}{W} - \frac{k-1}{W} \right) . \tag{8}$$

In particular this bound holds for the average cost of the LB algorithm:

$$\frac{k_{LB}-1}{W} C_A \left( \frac{k_{LB}-1}{W} \right) + \left( 1 - \frac{2}{W} - \frac{k_{LB}-1}{W} \right) C_B \left( 1 - \frac{2}{W} - \frac{k_{LB}-1}{W} \right) , \tag{9}$$

where as before $k_{LB}$ satisfies $C_A(k_{LB}/W) = C_B(1 - k_{LB}/W)$.

By appropriate choice of $C_A(.)$ and $C_B(.)$, we can ensure that the lower bound on the cost with the LB algorithm (Equation 9 evaluated with $k = k_{LB}$) is higher than the upper bound on the average cost incurred by the optimal algorithm (the minimum over $k$ of Equation 6). That is, the best possible average cost achieved by load balancing will be worse than the worst average cost that could arise through the optimal routing strategy. This establishes that LB does not engage in optimal routing.

**Example:** Let $C_A(x) = x^2$ and $C_B(x) = x$. Balancing the loads on $A$ and $B$ — setting $C_A(S(t)/W) = C_B(1 - S(t)/W)$ — results in $(S(t)/W)^2 = 1 - S(t)/W$, leading to $k_{LB}/W = \frac{\sqrt{5}-1}{2} = .618$. For $W = 1000$, the associated lower bound on average cost (Equation 9) is $(.618)^3 + (.998 - .618)^2 = .380$. On the other hand, with $C_A$ and $C_B$ given as above, Eq 6 is $(\frac{k+1}{W})^3 + (1 + \frac{2}{W} - \frac{k+1}{W})^2$. Differentiating with respect to $k$ and setting the result to zero leads to $\frac{k'}{W} = -\frac{1}{3} - \frac{1}{W} + \frac{\sqrt{28+48/W}}{6}$. For a window size of $W = 1000$, this yields $k'/W = .548$, a different result than $k_{LB}$. Plugging into Equation 6, the upper bound on the cost with $k'$ is $(.549)^3 + (1.002 - .549)^2 = .371$, which is less than .380.